\documentclass[runningheads]{llncs}
\usepackage{mwe} 
\usepackage[skip=0pt]{caption}
\usepackage{amssymb}
\usepackage{perpage}
\usepackage{graphicx}
\usepackage{wrapfig}
\usepackage{multirow}
\usepackage{graphicx}
\usepackage{lscape}
\usepackage{longtable}
\usepackage{float}
\usepackage{scrextend}
\usepackage{longtable}
\usepackage{bbding}

\MakePerPage{footnote}
\begin{document}
\title{Foundation Model Makes Clustering A Better Initialization For Cold-Start Active Learning}
\titlerunning{Foundation Model-Based Clustering For Cold-Start Active Learning}
%
\author{Han Yuan\inst{1} \and Chuan Hong\inst{2}}
\authorrunning{Han Yuan \& Chuan Hong}
%
\institute{Centre for Quantitative Medicine, Duke-NUS Medical School \and
Department of Biostatistics and Bioinformatics, Duke University}
\maketitle              
\begin{abstract}
Active learning selects the most informative samples from the unlabelled dataset to annotate in the context of a limited annotation budget. While numerous methods have been proposed for subsequent sample selection based on an initialized model, scant attention has been paid to the indispensable phase of active learning: selecting samples for model cold-start initialization. Most of the previous studies resort to random sampling or naive clustering. However, random sampling is prone to fluctuation, and naive clustering suffers from convergence speed, particularly when dealing with high-dimensional data such as imaging data. In this work, we propose to integrate foundation models with clustering methods to select samples for cold-start active learning initialization. Foundation models refer to those trained on massive datasets by the self-supervised paradigm and capable of generating informative and compacted embeddings for various downstream tasks. Leveraging these embeddings to replace raw features such as pixel values, clustering quickly converges and identifies better initial samples. For a comprehensive comparison, we included a classic ImageNet-supervised model to acquire embeddings. Experiments on two clinical tasks of image classification and segmentation demonstrated that foundation model-based clustering efficiently pinpointed informative initial samples, leading to models showcasing enhanced performance than the baseline methods. We envisage that this study provides an effective paradigm for future cold-start active learning.

\end{abstract}

\section{Introduction}
Based on large-scale samples as well as high-quality annotations, deep learning (DL) has emerged as the primary choice for dealing with high-dimensional medical images \cite{budd2021survey}. Although the data itself is no longer a challenge with the increasing adoption of electronic health records \cite{xie2022deep}, the gold-standard annotation by human experts is time-consuming and therefore becomes a bottleneck for developing DL for healthcare. Active learning is proposed to alleviate this issue and aims to select the most informative samples from the unlabelled dataset to annotate for model training \cite{wan2023survey} and various algorithms have been proposed, including uncertainty-based strategies, representative-based approaches, and hybrid methods \cite{gaillochet2023active}. However, most of these algorithms are designed for sample selection based on an initialized model, with scant attention given to the indispensable phase of active learning: selecting samples for cold-start model initialization. Conventionally, random sampling \cite{belharbi2021deep} and clustering \cite{clustering} are used for initializing cold-start active learning. However, random sampling is prone to fluctuation, and clustering encounters convergence challenges, especially when processing high-dimensional raw inputs such as pixel values \cite{ding2002adaptive}. To differentiate clustering based on distinct input features, we refer to clustering that utilizes raw inputs as 'naive clustering'. Apart from random sampling and naive clustering, prior studies have investigated weakly supervised methods for cold-start active learning initialization by using an auxiliary model for weak labels to identify initial samples for the target task \cite{wang2023alwod,zhao2021dsal}. In this study, weakly supervised initialization methods are not included due to the challenges associated with the weak label definition \cite{wang2023alwod}.

In this study, we propose to integrate foundation models with clustering to circumvent the fluctuation inherent in random sampling and address the convergence challenges encountered by naive clustering with high-dimensional inputs. Foundation models refer to those trained on massive datasets by the self-supervised paradigm and capable of generating low-dimensional yet information-rich embeddings, which are natural alternatives to original inputs of pixel values for clustering \cite{moor2023foundation}. Then samples closest to cluster medoids are selected as initial informative samples for model initialization. Our experiments, conducted on two clinical tasks of pneumothorax classification and segmentation, revealed that the initial samples selected by foundation model-based clustering exhibited superior performance compared to the baseline methods in both initialization and subsequent learning of cold-start active learning. We envisage that the proposed method provides an effective paradigm for future cold-start active learning.

\begin{figure}[!t]
\includegraphics[width=\textwidth]{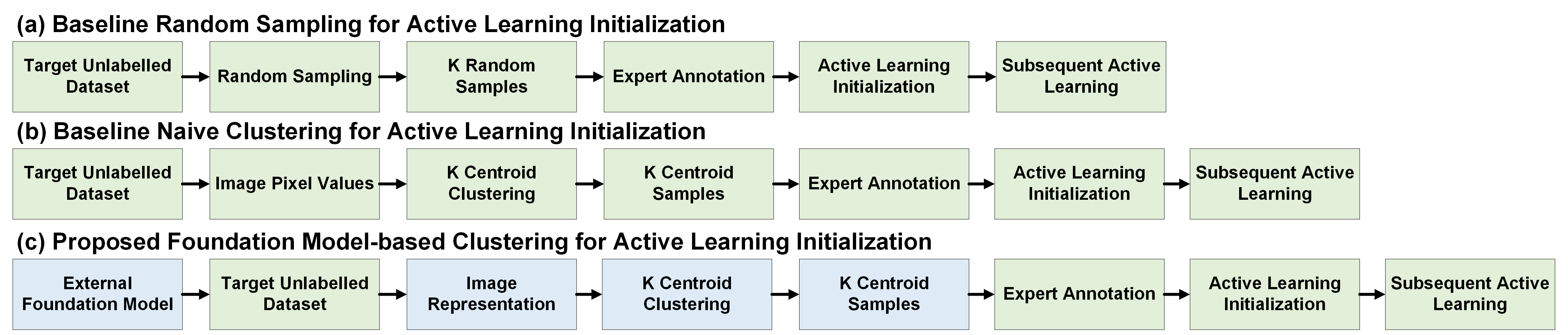}
\caption{Schematic comparison between the proposed and two baseline initialization approaches for cold-start active learning. The green blocks depict the shared modules while the blue blocks highlight the difference.}
\label{schematic}
\end{figure}

\section{Methods}
To harness the power of foundation models in improving cold-start active learning efficacy, we propose to integrate embeddings by foundation models with clustering methods for cold-start active learning initialization. In this section, we first present two baseline methods for initialization: random sampling and naive clustering. Then we introduce a subsequent learning strategy based on uncertainty to evaluate the impact of initialization on the the entire iteration cycle of cold-start active learning. Finally, we detail the foundation model-based clustering for initialization.

\subsection{Baseline Initialization Methods}
Traditional cold-start active learning applies either random sampling \cite{belharbi2021deep} or naive clustering \cite{clustering} for initialization. Subplots a and b in Figure \ref{schematic} depict the general steps of these two baseline methods. Random sampling randomly selects the pre-defined number of samples from the unlabelled dataset. Under a large sampling budget, random sampling selects diverse and representative samples while it suffers from instability when allocated a limited budget \cite{rudolph2023all}. To ensure comparability across different budgets, we condition that the sample set chosen by a low budget remains a subset of the sample set selected by a higher budget.

The other widely used initialization method in cold-start active learning is naive clustering \cite{clustering}, which partitions unlabelled samples into distinct clusters and selects samples closest to each cluster's center as the representative sample to comprise the samples for initialization. Whereas, when dealing with high-dimensional data such as images, the naive clustering method is prone to convergence difficulty \cite{ding2002adaptive}. Like random sampling, a strategy is designed to ensure the comparability of sample selection across various budgets. For a specific budget, clustering is implemented with increasing cluster numbers (i.e. cluster numbers 2, 3, 4, 5, ...), and the samples closest to cluster medoids are aggregated until the budget is fulfilled. K-means \cite{kmeans} is the default clustering method in this study.

To illustrate the influence of initialization on the entire iteration cycle of cold-start active learning, subsequent learning is conducted based on the classic uncertainty method. In a classification task, the most uncertain samples are those with predicted probabilities around the binarization threshold \cite{nguyen2022measure}. In a segmentation task, where the model outputs a probability vector for each pixel's likelihood of being included in the final segmentation area, uncertainty is computed as the mean uncertainty across all pixels \cite{yang2017suggestive}.

\subsection{Foundation Model-based Initialization}
As shown in Subplot c in Figure \ref{schematic}, the key distinction between foundation model-based clustering and naive clustering lies in the input features. Foundation models are capable of generating low-dimensional embeddings to mitigate the convergence difficulty caused by the raw pixel values in the process of clustering \cite{moor2023foundation,huang2023segment}. In this study, we explore three foundation models: TorchXRayVision (TXRV) \cite{TorchXRayVision}, CXR Foundation (CXRF) \cite{cxrfoundation}, and REMEDIS \cite{remedis}, which have proven to be effective in various downstream tasks \cite{soenksen2022integrated,rumberger2023actis,korot2023clinician}. TXRV is developed on 13 publicly available chest X-ray datasets with 728,004 thoracic disease-positive samples. CXRF is trained on 821,544 chest radiographs from India and the United States using a supervised contrastive approach. REMEDIS coveres five medical imaging domains: dermatology, mammography, digital pathology, fundus imaging, and chest radiographs through contrastive self-supervised learning. Through extensive training on large-scale datasets, these foundation models exhibit the capability to generate informative and condensed embeddings, serving as effective inputs for clustering. For TXRV \footnote{https://github.com/mlmed/torchxrayvision} and REMEDIS \footnote{https://github.com/google-research/medical-ai-research-foundations}, we extract embeddings from the last layer before the classification head. For CXRF, the concrete architecture is not publicly accessible and the embeddings are acquired from the granted online interface \footnote{https://github.com/Google-Health/imaging-research/tree/master/cxr-foundation}. Besides, we extract embeddings from the ImageNet-supervised DenseNet-121 \cite{densenet} as a baseline to investigate whether the embeddings from the latest foundation models could surpass those from a previous canonical model. After model initialization, the uncertainty-based subsequent learning is implemented as the baseline initialization.

\section{Experiments}
We demonstrated the effectiveness of foundation model-based clustering for cold-start active learning initialization using the real-world dataset of ChestX-Det \cite{liu2020chestx}. The ChestX-Det dataset consists of 611 healthy images and 189 pneumothorax-positive images. Apart from the binary labels of pneumothorax, each positive sample contains pixel-level annotations of lesion areas. Therefore, we conducted two tasks of pneumothorax classification and pneumothorax segmentation to present the superiority and generalizability of foundation model-based clustering. For pneumothorax classification, We randomly split the whole dataset into three parts at 70: 10: 20 for training set, validation set, and test set. For pneumothorax segmentation, the pneumothorax-negative samples from the previous sets were excluded and the positive samples were retained. For each initialization method in either classification or segmentation task, it selected samples from the training set under different budgets of 20, 40, 60, 80, and 100. Then the selected samples would be supplemented with annotations for model initialization while the other samples remained unlabeled. To further illustrate the impact of foundation model-based initialization on the complete iteration cycle of cold-start active learning, we applied the uncertainty method for subsequent learning of the model initialized by 20 samples. Throughout each iteration of subsequent learning, a budget of 20 samples was allocated for model updates.

For pneumothorax classification, we implemented the classifier with a VGG-11 model \cite{vgg}. This choice was motivated by the previous study that in limited training data settings, neural networks with a lightweight architecture demonstrated comparable or even superior performance to a complex one \cite{ccayir2023patch,ma2024segment}. Following the training settings in a previous study \cite{jiang2022harmofl}, we used the Stochastic Gradient Descent (SGD) \cite{sgd} optimizer with an initial learning rate of 1e-3 and a momentum of 0.9. Model training was conducted in batches of 10 images, using weighted cross-entropy as the loss function to counterbalance the predominance of negative samples \cite{imbalance}. The training was set as 100 epochs with a scheduler to halve the current learning rate if no improvement was observed on the validation set over 10 consecutive epochs. After the training, the classifier's performance was evaluated on the test set. Considering the data imbalance, we applied the area under the precision recall curve (AUPRC) and F1 Score (F1) to evaluate the classifiers trained on samples selected by different initialization methods \cite{imbalance1,imbalance2}. Higher AUPRC and F1 values indicate better performance. For each metric, standard errors (SE) were calculated using the nonparametric bootstrap of 100 times \cite{bootstrap}.

For pneumothorax segmentation, we developed the segmenter based on U-Net architecture \cite{ronneberger2015u} with a backbone of VGG-11 \cite{vgg}. Except for maintaining the initial learning rate at 1e-2 and utilizing the Dice Similarity Coefficient (DSC) \cite{dice} as the loss function\cite{eelbode2020optimization}, all other training settings remained consistent with those employed in the classification task. After the training, the segmenter's performance was evaluated on the test set. Similar to a prior study \cite{yuan2023leveraging}, we applied DSC and Hausdorff Distance (HD) \cite{Hausdorff} to evaluate the segmenters. Higher DSC values indicate better performance, while lower HD values indicate better results. SE for each metric was computed using the nonparametric bootstrap of 100 times \cite{bootstrap}. To ensure experiment reproducibility, the random seeds were set to 2024 and the code is available at GitHub \footnote{https://github.com/Han-Yuan-Med/foundation-model}.

\begin{table}[!t]
\centering
\caption{Classification performance of VGG-11 trained on samples selected by different initialization methods. The evaluation metrics accompanied by their respective SE on the test set are presented. Higher AUPRC or F1 values indicate better performance}
\label{cls-result}
\begin{tabular}{ccccc}
\hline
Initialization Budget                       & \multicolumn{2}{c}{Method}                    & AUPRC                  & F1                     \\ \hline
\multirow{6}{*}{20 samples}  & \multicolumn{2}{c}{Random Sampling}           & 0.389 (0.094)          & 0.447 (0.100)          \\ 
                             & \multirow{5}{*}{Clustering} & Naive           & 0.313 (0.056)          & 0.390 (0.051)          \\
                             &                             & ImageNet        & 0.545 (0.076)          & 0.482 (0.050)          \\
                             &                             & TXRV & \textbf{0.557 (0.082)} & 0.524 (0.071)          \\
                             &                             & REMEDIS         & 0.420 (0.068)          & 0.412 (0.047)          \\
                             &                             & CXRF  & 0.506 (0.083)          & \textbf{0.554 (0.059)} \\ \hline
\multirow{6}{*}{40 samples}  & \multicolumn{2}{c}{Random Sampling}           & 0.517 (0.113)          & 0.508 (0.071)          \\ 
                             & \multirow{5}{*}{Clustering} & Naive           & 0.386 (0.061)          & 0.462 (0.052)          \\
                             &                             & ImageNet        & 0.507 (0.093)          & 0.585 (0.052)          \\
                             &                             & TXRV & 0.617 (0.091)          & \textbf{0.608 (0.047)} \\
                             &                             & REMEDIS         & 0.371 (0.062)          & 0.463 (0.061)          \\
                             &                             & CXRF  & \textbf{0.640 (0.071)} & 0.536 (0.064)          \\ \hline
\multirow{6}{*}{60 samples}  & \multicolumn{2}{c}{Random Sampling}           & 0.585 (0.098)          & 0.554 (0.080)          \\ 
                             & \multirow{5}{*}{Clustering} & Naive           & 0.513 (0.083)          & 0.538 (0.053)          \\
                             &                             & ImageNet        & \textbf{0.658 (0.089)} & \textbf{0.620 (0.053)} \\
                             &                             & TXRV & 0.532 (0.080)          & 0.585 (0.045)          \\
                             &                             & REMEDIS         & 0.524 (0.079)          & 0.529 (0.054)          \\
                             &                             & CXRF  & 0.624 (0.079)          & 0.566 (0.058)          \\ \hline
\multirow{6}{*}{80 samples}  & \multicolumn{2}{c}{Random Sampling}           & 0.649 (0.106)          & 0.596 (0.068)          \\ 
                             & \multirow{5}{*}{Clustering} & Naive           & 0.553 (0.082)          & 0.589 (0.057)          \\
                             &                             & ImageNet        & 0.660 (0.079)          & 0.607 (0.049)          \\
                             &                             & TXRV & 0.522 (0.077)          & 0.512 (0.064)          \\
                             &                             & REMEDIS         & 0.572 (0.075)          & 0.574 (0.066)          \\
                             &                             & CXRF  & \textbf{0.750 (0.061)} & \textbf{0.675 (0.056)} \\ \hline
\multirow{6}{*}{100 samples} & \multicolumn{2}{c}{Random Sampling}           & 0.649 (0.091)          & 0.602 (0.063)          \\ 
                             & \multirow{5}{*}{Clustering} & Naive      & 0.592 (0.076)          & 0.582 (0.049)          \\
                             &                             & ImageNet        & 0.721 (0.082)          & \textbf{0.707 (0.052)} \\
                             &                             & TXRV & 0.646 (0.076)          & 0.598 (0.050)          \\
                             &                             & REMEDIS         & 0.625 (0.071)          & 0.659 (0.058)          \\
                             &                             & CXRF  & \textbf{0.762 (0.064)} & 0.621 (0.052)          \\ \hline
\end{tabular}
\end{table}

\begin{table}[!t]
\centering
\caption{Classification performance of initialized VGG-11 trained by the uncertainty-based subsequent learning (Budget: 20 samples in each iteration). The evaluation metrics accompanied by their respective SE on the test set are presented. Higher AUPRC or F1 values indicate better performance}
\label{cls-subsequent}
\begin{tabular}{ccccc}
\hline
Subsequent Budget                       & \multicolumn{2}{c}{Method}                    & AUPRC                  & F1                     \\ \hline
\multirow{6}{*}{20 samples}  & \multicolumn{2}{c}{Random sampling} & 0.501   (0.115) & 0.484 (0.103) \\
                            & \multirow{5}{*}{Clustering} & Naive           & 0.400   (0.071)          & 0.448   (0.044)          \\
                            &                             & ImageNet        & 0.557   (0.078)          & 0.548   (0.049)          \\
                            &                             & TXRV & 0.500 (0.070)            & 0.577 (0.057)            \\
                            &                             & REMEDIS         & 0.496   (0.081)          & 0.526 (0.055)            \\
                            &                             & CXRF  & \textbf{0.631 (0.066)}   & \textbf{0.579 (0.057)}   \\ \hline
\multirow{6}{*}{40 samples} & \multicolumn{2}{c}{Random sampling}           & 0.562 (0.097)            & 0.559 (0.090)            \\
                            & \multirow{5}{*}{Clustering} & Naive           & 0.560   (0.077)          & 0.580   (0.054)          \\
                            &                             & ImageNet        & 0.714   (0.064)          & \textbf{0.646   (0.064)} \\
                            &                             & TXRV & 0.619 (0.077)            & 0.595 (0.055)            \\
                            &                             & REMEDIS         & 0.608 (0.072)            & 0.543 (0.070)            \\
                            &                             & CXRF  & \textbf{0.722 (0.063)}   & 0.620 (0.051)            \\ \hline
\multirow{6}{*}{60 samples} & \multicolumn{2}{c}{Random sampling}           & 0.660 (0.092)            & 0.630 (0.071)            \\
                            & \multirow{5}{*}{Clustering} & Naive           & 0.485   (0.070)          & 0.561   (0.047)          \\
                            &                             & ImageNet        & 0.697   (0.079)          & \textbf{0.659   (0.062)} \\
                            &                             & TXRV & \textbf{0.735 (0.061)}   & 0.594 (0.065)            \\
                            &                             & REMEDIS         & 0.589 (0.075)            & 0.574 (0.059)            \\
                            &                             & CXRF  & 0.727 (0.074)            & 0.639 (0.048)            \\ \hline
\multirow{6}{*}{80 samples} & \multicolumn{2}{c}{Random sampling}           & 0.640 (0.111)            & 0.578 (0.099)            \\
                            & \multirow{5}{*}{Clustering} & Naive           & 0.544   (0.076)          & 0.511   (0.059)          \\
                            &                             & ImageNet        & \textbf{0.808   (0.053)} & \textbf{0.699   (0.049)} \\
                            &                             & TXRV & 0.765 (0.065)            & 0.688 (0.054)            \\
                            &                             & REMEDIS         & 0.476 (0.080)            & 0.489 (0.058)            \\
                            &                             & CXRF  & 0.768 (0.056)            & 0.694 (0.059)            \\ \hline
\multirow{6}{*}{100 samples} & \multicolumn{2}{c}{Random sampling} & 0.696 (0.106)   & 0.624 (0.083) \\
                            & \multirow{5}{*}{Clustering} & Naive           & 0.566   (0.087)          & 0.450   (0.061)          \\
                            &                             & ImageNet        & 0.691   (0.053)          & 0.598   (0.060)          \\
                            &                             & TXRV & \textbf{0.771 (0.059)}   & \textbf{0.690 (0.056)}   \\
                            &                             & REMEDIS         & 0.739 (0.065)            & 0.576 (0.055)            \\
                            &                             & CXRF  & 0.771 (0.069)            & 0.685 (0.050)   \\ \hline
\end{tabular}
\end{table}

\begin{table}[!t]
\centering
\caption{Segmentation performance of U-Net with VGG-11 backbone trained on samples selected by different initialization methods. The evaluation metrics accompanied by their respective SE on the test set are presented. Higher DSC or lower HD values indicate better performance.}
\label{seg-result}
\begin{tabular}{ccccc}
\hline
Initialization Budget                       & \multicolumn{2}{c}{Method}                    & DSC                    & HD                     \\ \hline
\multirow{6}{*}{20 samples}  & \multicolumn{2}{c}{Random Sampling}           & 0.161 (0.051)          & 4.976 (0.124)          \\ 
                             & \multirow{5}{*}{Clustering} & Naive           & 0.032 (0.004)          & 13.134 (0.111)         \\
                             &                             & ImageNet        & 0.141 (0.031)          & \textbf{4.267 (0.232)} \\
                             &                             & TXRV & \textbf{0.244 (0.031)} & 4.607 (0.228)          \\
                             &                             & REMEDIS         & 0.120 (0.024)          & 7.143 (0.432)          \\
                             &                             & CXRF  & 0.109 (0.022)          & 5.762 (0.280)          \\ \hline
\multirow{6}{*}{40 samples}  & \multicolumn{2}{c}{Random Sampling}           & 0.265 (0.052)          & 4.497 (0.098)          \\ 
                             & \multirow{5}{*}{Clustering} & Naive           & 0.275 (0.038)          & 4.624 (0.199)          \\
                             &                             & ImageNet        & 0.207 (0.038)          & 4.661 (0.189)          \\
                             &                             & TXRV & \textbf{0.291 (0.042)} & 3.961 (0.201)          \\
                             &                             & REMEDIS         & 0.278 (0.047)          & \textbf{3.956 (0.216)} \\
                             &                             & CXRF  & 0.291 (0.047)          & 4.309 (0.174)          \\ \hline
\multirow{6}{*}{60 samples}  & \multicolumn{2}{c}{Random Sampling}           & 0.290 (0.046)          & 4.683 (0.098)          \\ 
                             & \multirow{5}{*}{Clustering} & Naive           & 0.384 (0.034)          & 4.188 (0.173)          \\
                             &                             & ImageNet        & 0.244 (0.041)          & 4.447 (0.187)          \\
                             &                             & TXRV & \textbf{0.398 (0.038)} & 4.908 (0.226)          \\
                             &                             & REMEDIS         & 0.297 (0.041)          & 4.563 (0.174)          \\
                             &                             & CXRF  & 0.246 (0.039)          & \textbf{4.016 (0.206)} \\ \hline
\multirow{6}{*}{80 samples}  & \multicolumn{2}{c}{Random Sampling}           & 0.330 (0.057)          & 4.285 (0.111)          \\ 
                             & \multirow{5}{*}{Clustering} & Naive           & 0.296 (0.047)          & 4.287 (0.197)          \\
                             &                             & ImageNet        & 0.306 (0.036)          & 4.759 (0.195)          \\
                             &                             & TXRV & 0.340 (0.048)          & 3.818 (0.198)          \\
                             &                             & REMEDIS         & \textbf{0.392 (0.040)} & \textbf{3.795 (0.174)} \\
                             &                             & CXRF  & 0.327 (0.043)          & 4.032 (0.209)          \\ \hline
\multirow{6}{*}{100 samples} & \multicolumn{2}{c}{Random Sampling}           & 0.353 (0.051)          & 4.525 (0.085)          \\ 
                             & \multirow{5}{*}{Clustering} & Naive      & 0.375 (0.041)          & 4.286 (0.169)          \\
                             &                             & ImageNet        & 0.379 (0.033)          & 4.518 (0.157)          \\
                             &                             & TXRV & 0.381 (0.040)          & \textbf{4.003 (0.220)} \\
                             &                             & REMEDIS         & \textbf{0.392 (0.040)} & 4.836 (0.201)          \\
                             &                             & CXRF  & 0.364 (0.047)          & 4.046 (0.182)         \\ \hline
\end{tabular}
\end{table}

\begin{table}[!t]
\centering
\caption{Segmentation performance of initialized U-Net with VGG-11 backbone trained by the uncertainty-based subsequent learning (Budget: 20 samples in each iteration). The evaluation metrics accompanied by their respective SE on the test set are presented. Higher DSC or lower HD values indicate better performance.}
\label{seg-subsequent}
\begin{tabular}{ccccc}
\hline
Subsequent Budget                       & \multicolumn{2}{c}{Method}                    & AUPRC                  & F1                     \\ \hline
\multirow{6}{*}{20 samples}  & \multicolumn{2}{l}{Random sampling} & 0.270   (0.061) & 4.639 (0.527) \\
                            & \multirow{5}{*}{Clustering} & Naive           & 0.225   (0.042)          & 4.364   (0.172)          \\
                            &                             & ImageNet        & 0.281   (0.048)          & 4.364   (0.180)          \\
                            &                             & TXRV & \textbf{0.284 (0.039)}   & 4.693 (0.239)            \\
                            &                             & REMEDIS         & 0.241   (0.027)          & 5.214 (0.234)            \\
                            &                             & CXRF  & 0.248 (0.042)            & \textbf{3.922 (0.184)}   \\ \hline
\multirow{6}{*}{40 samples} & \multicolumn{2}{c}{Random sampling}           & 0.303 (0.053)            & 4.458 (0.374)            \\
                            & \multirow{5}{*}{Clustering} & Naive           & \textbf{0.312   (0.040)} & 4.326   (0.208)          \\
                            &                             & ImageNet        & 0.280   (0.049)          & 4.051   (0.202)          \\
                            &                             & TXRV & 0.305 (0.050)            & \textbf{3.958 (0.217)}   \\
                            &                             & REMEDIS         & 0.297 (0.035)            & 5.364 (0.236)            \\
                            &                             & CXRF  & 0.308 (0.036)            & 4.744 (0.178)            \\ \hline
\multirow{6}{*}{60 samples} & \multicolumn{2}{c}{Random sampling}           & 0.317 (0.047)            & 4.416 (0.514)            \\
                            & \multirow{5}{*}{Clustering} & Naive           & \textbf{0.369   (0.039)} & \textbf{3.930   (0.190)} \\
                            &                             & ImageNet        & 0.293   (0.033)          & 5.292   (0.216)          \\
                            &                             & TXRV & 0.312 (0.034)            & 5.087 (0.230)            \\
                            &                             & REMEDIS         & 0.283 (0.042)            & 5.179 (0.169)            \\
                            &                             & CXRF  & 0.315 (0.035)            & 4.558 (0.199)            \\ \hline
\multirow{6}{*}{80 samples} & \multicolumn{2}{c}{Random sampling}           & 0.337 (0.051)            & 4.390 (0.452)            \\
                            & \multirow{5}{*}{Clustering} & Naive           & 0.346   (0.035)          & 4.452 (0.162)            \\
                            &                             & ImageNet        & 0.329   (0.039)          & 4.372   (0.195)          \\
                            &                             & TXRV & 0.349 (0.031)            & 5.320 (0.158)            \\
                            &                             & REMEDIS         & \textbf{0.372 (0.041)}   & \textbf{3.841 (0.211)}   \\
                            &                             & CXRF  & 0.282 (0.039)            & 5.063 (0.214)            \\ \hline
\multirow{6}{*}{100 samples} & \multicolumn{2}{c}{Random sampling} & 0.356 (0.052)   & 4.351 (0.530) \\
                            & \multirow{5}{*}{Clustering} & Naive           & 0.400   (0.044)          & 4.022   (0.176)          \\
                            &                             & ImageNet        & 0.389   (0.042)          & 4.155   (0.172)          \\
                            &                             & TXRV & 0.360 (0.039)            & \textbf{3.970 (0.168)}   \\
                            &                             & REMEDIS         & 0.317 (0.032)            & 5.068 (0.260)            \\
                            &                             & CXRF  & \textbf{0.391 (0.039)}   & 4.293 (0.204)           \\ \hline
\end{tabular}
\end{table}

\section{Results}
This section presents the test results of the classifier and segmenter initialized by samples from different methods, accompanied by the respective SE enclosed within parentheses. Table \ref{cls-result} showed the classification performance of VGG-11 developed on samples selected by different initialization methods. In terms of AUPRC, CXRF-based clustering significantly outperformed baseline methods and other foundation model-based clustering methods, showcasing the best performance in budgets of 40, 80, and 100. At budget levels of 20 and 60, the highest AUPRC values were attained by TXRV-based clustering and ImageNet-based clustering, respectively. Regarding F1 scores, both ImageNet-based clustering and CXRF-based clustering achieved the highest performance in two out of five scenarios. Assessed by AUPRC or F1, REMEDIS-based clustering showed inferior performance compared to at least one of the baseline methods in four out of five scenarios, underscoring the disparity in foundation models' ability to generate informative embeddings for pneumothorax classifier initialization. Meanwhile, random sampling remained a powerful method, particularly evident in relatively high budgets of 60, 80, and 100 samples, securing the second-highest ranking in terms of AUPRC while it also presented the highest instability across all scenarios and evaluation metrics.

Utilizing the classifier initialized by 20 samples, we employed the uncertainty method for the subsequent active learning. In each iteration of the subsequent active learning, 20 predominantly uncertain samples were annotated from the unlabeled set and combined with the previously labeled samples for model updates. In the subsequent learning stage, regarding AUPRC, both classifiers initialized by CXRF-based and TXRV-based clustering exhibited optimal performance in two out of five scenarios. However, when evaluated by F1, ImageNet-based clustering yielded the best model in three out of five scenarios and in the remaining two scenarios, CXRF-based and TXRV-based clustering produced the best models.

Table \ref{seg-result} displays the segmentation performance of U-Net with VGG-11 backbone developed on samples chosen by different initialization methods. In terms of DSC, TXRV-based clustering attained the highest performance in three out of five scenarios and REMEDIS-based clustering achieved the best performance in the other two scenarios. Measured by HD, TXRV, REMEDIS, and CXRF-based clustering achieved the best performance in at least one scenario. Consistent with the classification task, random sampling showed the highest SE of DSC in all scenarios. Nevertheless, it only achieved the second-highest DSC or HD in one scenario, which was not comparable to its success in the classification task.

In the subsequent learning stage of pneumothorax segmentation, the segmenter initialized by TXRV-based clustering achieved the highest DSC in three out of five scenarios, while the segmenter initialized by REMEDIS-based clustering excelled in the remaining two scenarios. Regarding HD, the segmenter initialized by REMEDIS-based clustering obtained the two best values, while the remaining three best values were achieved by the segmenter initialized by ImageNet, TXRV, and CXRF-based clustering, respectively.

\section{Discussion}
This study presents an effective method for cold-start active learning initialization by integrating foundation models with clustering. Leveraging the informative embeddings generated by foundation models, the clustering selected better samples and contributed to models with superior performance than the baseline methods in both stages of initialization and subsequent learning \cite{yang2022actively}. 

Performance divergence was observed among the three foundation model-based clustering methods, with REMEDIS-based clustering exhibiting comparatively lower performance. In contrast to the other two models specialized in chest radiographs, REMEDIS was designed as a versatile model capable of handling five medical imaging modalities, exemplifying a phenomenon where a generalist model may not achieve comparable performance to a specialist in healthcare. Moreover, ImageNet-based embeddings showed comparable performance to three foundation model-based clustering methods in the classification task while relatively bad performance in the segmentation task, indicating the ImageNet-based embeddings are capable of selecting initial samples for simple classification rather than complex segmentation. Besides, Table \ref{cls-result} depicts multiple performance drops with increased budgets in cold-start active learning initialization, and Table \ref{cls-subsequent} and \ref{seg-subsequent} shows several similar drops when the model was updated with higher budgets in subsequent active learning. We suggested that the fluctuation of evaluation metrics could stem from three potential sources: the model convergence affected by SGD, the model optima selected by the limited validation samples, and the evaluation inconsistency due to the small test dataset. To alleviate the instability problem, a larger dataset and multiple simulations are necessary to stabilize parameter estimation and substantiate the positive correlation between model performance and annotation budgets \cite{samplesize,byrd2012sample}.

\section{Conclusion}
Based on low-dimensional embeddings generated by foundation models, clustering selected more informative samples for initializing cold-start active learning, leading to models showcasing enhanced performance than the baselines. We hope this study provides an effective paradigm for future cold-start active learning.

\bibliographystyle{splncs04}
\bibliography{citation}

\end{document}